\pdfoutput=1
\documentclass[11pt]{article}

\usepackage[]{emnlp2021}

\usepackage{times}
\usepackage{latexsym}
\usepackage{arydshln}

\usepackage[T1]{fontenc}
\usepackage[utf8]{inputenc}

\usepackage{microtype}

\usepackage{algorithmic}
\usepackage{latexsym}

\usepackage[ruled,vlined]{algorithm2e}
\usepackage{amsmath}
\usepackage{graphicx}
\usepackage{booktabs}
\usepackage{comment}

\title{Unsupervised Contextualized Document Representation}

\author{Ankur Gupta \\
  Indian Institute of Technology, Kanpur \\
  \texttt{ankugupt@iitk.ac.in} \\\And
  Vivek Gupta \\
  School of Computing, University of Utah \\
  \texttt{vgupta@cs.utah.edu} \\}

\begin{document}
\maketitle
\begin{abstract}

Several NLP tasks need the effective representation of text documents. \citealp{arora2017asimple} demonstrate that simple weighted averaging of word vectors frequently outperforms neural models. SCDV \cite{mekala-etal-2017-scdv} further extends this from sentences to documents by employing soft and sparse clustering over pre-computed word vectors. However, both techniques ignore the polysemy and contextual character of words. In this paper, we address this issue by proposing SCDV+BERT(ctxd), a simple and effective unsupervised representation that combines contextualized BERT \cite{devlin-etal-2019-bert} based word embedding for word sense disambiguation with SCDV soft clustering approach. We show that our embeddings outperform original SCDV, pre-train BERT, and several other baselines on many classification datasets. We also demonstrate our embeddings effectiveness on other tasks, such as concept matching and sentence similarity. In addition, we show that SCDV+BERT(ctxd) outperforms fine-tune BERT and different embedding approaches in scenarios with limited data and only few shots examples.
\end{abstract}

\section{Introduction}

The semantics of a document is highly dependent on the constituent words, and words can have different meaning in different contexts. Approaches such as \citealp{socher-etal-2013-recursive}; \citealp{Liu2015LearningCW}; \citealp{10.5555/3044805.3045025}; \citealp{ling-etal-2015-two} go beyond words to capture the semantic meaning of sentences but are restricted to perceiving the meaning of a single sentence, thus reducing its expressive power. 

A simple weighted average of the individual word embeddings doesn't account for word ordering and long-distance semantic relationships. \citealp{gupta-etal-2016-product,mekala-etal-2017-scdv} proposed clustering-based technique with tf-idf weighting to form sparse composite document vector (SCDV), thus extending the simple averaging approach beyond a single sentence. Recently, \citealp{gupta2020multisense} introduced SCDV-MS, which highlights how multi-sense word embedding resolves cluster disambiguation, which improves embedding performance, further enhancing SCDV. \citealp{gupta2020psif} (PSIF) additionally demonstrates that a sparsity constraint in clustering can be advantageous.

Modern contextualized representations such as \citealp{devlin-etal-2019-bert} can capture the exact meaning of a word based on the surrounding context, which can automatically disambiguate the meaning of words in a corpus-based on its interpretations. Previous approaches for document representation, as discussed above, ignore these contextualized representation benefits. Therefore, in this work, we propose a new approach (SCDV+BERT(ctxd), which leverages a combination of clustering techniques for word-sense disambiguation and combines it with the expressibility of SCDV partition averaging method for creating better document representation. We contextualized the corpus by using the word sense disambiguation power contextualized pre-train BERT embedding \footnote{For simplicity, we picked the BERT model. Others language models, such as RoBERTa, should work equally well.}. Contextual embedding from pre-train BERT disambiguate the occurrence of the same word with different context. 

The contextualized corpus is then converted into document embeddings utilizing pre-train BERT word embeddings as static word vectors using the SCDV partition averaging technique. We show that our unsupervised embeddings SCDV+BERT(ctxd) outperform existing techniques on several classification datasets in supervised, semi-supervised, and few-shot settings. We also demonstrate performance improvement in non-classification tasks such as concept matching and sentence similarity using our representation. The datasets along with the associated scripts, can be located at \url{https://github.com/vgupta123/contextualize_scdv}.

\section{Proposed Algorithm}
\label{sec:proposed_algorithm}

\begin{figure*}[!htbp]
\centering
\includegraphics[keepaspectratio, width=0.95\textwidth]{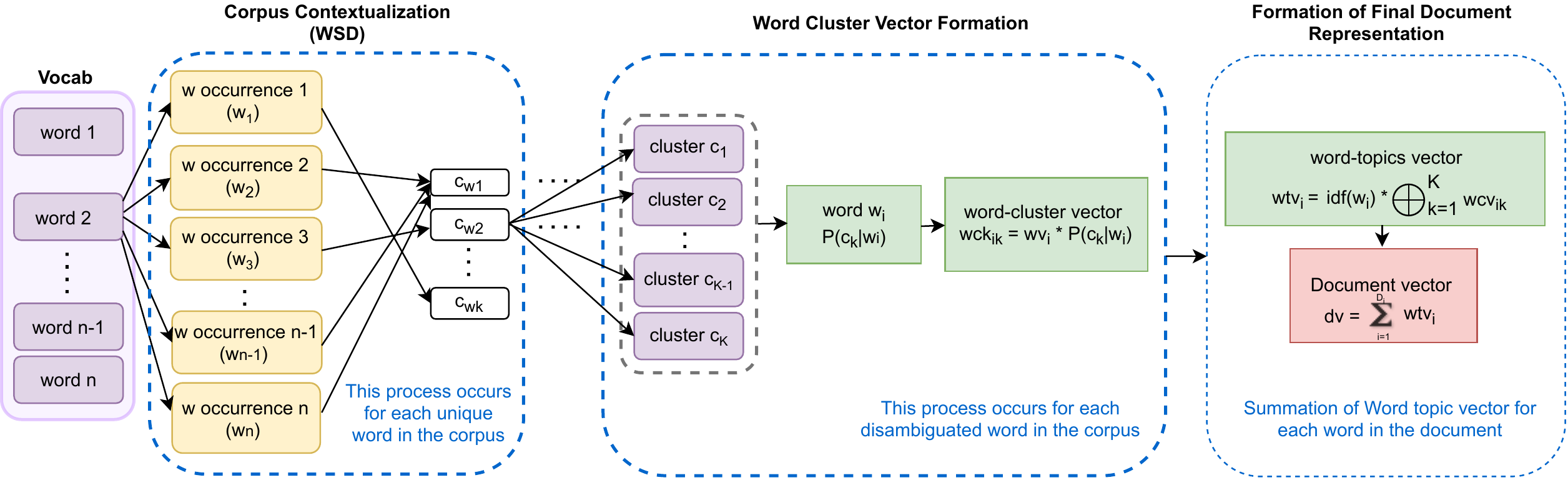}
\vspace{-0.25em}
\caption{\small High level flowchart of our document representation approach a.k.a SCDV + BERT(cxtd)}
\label{tab:flowchart}
\vspace{-0.5em}
\end{figure*}

Our algorithm is similar to SCDV algorithm, but uses word sense disambiguated contextualize word vectors obtained from pre-trained BERT Embedding as static word vectors. Below are details of the primary steps involved.

\paragraph{Corpus Contextualization (WSD):}

Our objective here is to disambiguate different occurrences of a word in a corpus document with its separate interpretation. For example, the word `\textit{bank}' in \emph{``He went to `\textit{bank}' for withdrawing money"} and \emph{``I went to a river `\textit{bank}' during summer holiday"} has different meanings based on it's used context. Given a word $w$ and all its occurrence in the corpus documents as $w_{1}$, $w_{2}$, $\ldots$ , $w_{n}$, for each $w_{i}$ , we find its contextualized embedding representation $b_{w_{i}}$ using transformer based pre-trained language model such as BERT \cite{devlin-etal-2019-bert}.

Taking inspiration from \citealp{mekala-etal-2020-meta}, we treat the word disambiguation problem as a local clustering problem of contextualizing word vectors. For our case, we also cluster the contextualize word embedding $b_{w_{i}}$, obtained via the pre-train BERT model from the last step. We use the efficient K-means algorithm to cluster all $b_{w_{i}}$ into $k$ clusters, where $k$ represents the total possible interpretations of word $w$ in all the documents of the corpus.\footnote{K-means is computationally more efficient than other clustering approaches. Other clustering algorithm also works.} We use the cosine distance between the contextualize word representation as our clustering metric. \footnote{L2 on normalized vectors is same as the cosine distance.} The value of $k$ i.e., cluster numbers, is decided using a similarity threshold ($\tau$), which is a hyperparameter and a dataset property, usually set using the heuristic described in \cite{mekala-etal-2020-meta}.  

Let $c_{w_1}$, $c_{w_2}$, $\ldots$ , $c_{w_k}$ be the $k$ cluster centroids obtained after the K-means clustering for the word $w$. We treat these $k$ centroid representation as our polysemous word representations highlighting the $k$ sense of the words $w$. After clustering for each occurrence of word $w$ in a corpus, we perform contextualised word sense disambiguation using cosine similarity between it's BERT representation and our centroid embedding i.e. $c_{w_1}$, $c_{w_2}$, $\ldots$ , $c_{w_k}$ to discover the closest cluster centroid $j$ a.k.a nearest neighbour (j), the word sense for that occurrence. 

Finally, we assign this nearest neighbor cluster centroids embedding $c_{w_{j}}$ as the contextualized disambiguated word embedding for that word $w$ occurrence. We repeat this procedure for all the occurrences of word $w$ to obtain final sense disambiguated contextualized word embedding. Each of these contextualized embeddings of word $w$ act as our distant sense-disambiguated word vectors.

\begin{table*}[!htbp]
\vspace{0.25em}
\small
\centering
\begin{tabular}{ c|c|c|c|c|c|c } 
 \toprule
 \bf Embedding & \bf Amazon & \bf BBCSport & \bf Twitter & \bf Classic & \bf Recipe-L & \bf 20NG \\
 \midrule
 SIF(Glove) & 94.1$_{(0.2)}$ & 97.3$_{(1.2)}$ & 57.8$_{(2.5)}$ & 92.7$_{(0.9)}$ & 71.1$_{(0.5)}$ & 72.3$_{(0.11)}$ \\
%  Word2Vec (Averaged) & 94.0$_{(0.5)}$ & 97.3$_{(0.9)}$ & 72.0$_{(1.5)}$ & 95.2$_{(0.4)}$ & 74.9$_{(0.5)}$ & 71.7 \\
 %PV-DBOW & 89.2$_{(0.3)}$ & 97.2$_{(0.7)}$ & 67.8$_{(0.4)}$ & 97.0$_{(0.3)}$ & 73.1$_{(0.5)}$ & 71.0 \\
 PV-DM & 88.6$_{(0.4)}$ & 97.9$_{(1.3)}$ & 67.3$_{(0.3)}$ & 96.5$_{(0.7)}$ & 71.1$_{(0.4)}$ & 74.0$_{(0.11)}$ \\
 Doc2VecC & 91.2$_{(0.5)}$ & 90.5$_{(1.7)}$ & 71.0$_{(0.4)}$ & 96.6$_{(0.4)}$ & 76.1$_{(0.4)}$ & 78.2$_{(0.11)}$ \\
% KNN-WMD & 92.6$_{(0.3)}$ & 95.4$_{(1.2)}$ & 71.3$_{(0.6)}$ & 97.2$_{(0.1)}$ & 71.4$_{(0.5)}$ & 73.2 \\
 Word2Vec (idf-weighted) & 94.00$_{(0.45)}$ & 97.30$_{(0.67)}$ & 72.00$_{(0.36)}$ & 95.20$_{(0.44)}$ & 74.90$_{(0.89)}$ & 81.70$_{(0.22)}$ \\ 
 BERT(pr) & 91.04$_{(0.27)}$ & 99.12$_{(0.66)}$ & 66.63$_{(0.22)}$ & 95.63$_{(0.36)}$ & 68.44$_{(0.07)}$ & 64.81$_{(0.17)}$ \\ 
 SCDV + Word2Vec & 93.90$_{(0.40 )}$& 98.81$_{(0.60)}$& 74.20$_{(0.40)}$& 96.90$_{(0.10)}$& 78.50$_{(0.50)}$&  84.90$_{(0.13)}$ \\ 
 SCDV + BERT(weight-avg) & 94.62$_{(0.21)}$ & 97.29$_{(0.56)}$ & 72.98$_{(0.24)}$ & 96.54$_{(0.61)}$ & 78.13$_{(0.15)}$ & 84.90$_{(0.13)}$ \\ \hdashline
 SCDV + BERT(ctxd) & 94.16$_{(0.31)}$ & 99.58$_{(0.41)}$ & 75.98$_{(0.36)}$ & 97.84$_{(0.40)}$ & 80.71$_{(0.19)}$ & 86.12$_{(0.11)}$ \\
SCDV + BERT(ctxd)  & \bf 95.88$_{(0.34)}$ &\bf 99.60$_{(0.59)}$ & \bf 77.03$_{(0.27)}$ & \bf 99.01$_{(0.41)}$ & 80.74$_{(0.15)}$ &\bf 86.94$_{(0.11)}$ \\ 
+ Anisotropy & & & & & &\\
 \midrule
  BERT (fine-tune) & 94.60$_{(0.19)}$ & \bf 99.67$_{(0.51)}$ & 73.13$_{(0.31)}$ &\bf 98.67$_{(0.56)}$ &\bf 81.13$_{(0.21)}$ &\bf 86.91$_{(0.28)}$ \\ 
 \bottomrule
\end{tabular}
\vspace{-0.5em}
\caption{\small Embedding performance with complete training i.e. full data setting. Bold represents best performance.  Reported number are means performance and subscript brackets number x $_{(x)}$ represent the standard deviation over $5$ random runs. Baselines are taken from \citealp{gupta2020psif}. We use $k = 6$ for the ainsotropic adjustment.}
\label{tab:my-table}
\vspace{-1.5em}
\end{table*}

\paragraph{Document Representation (SCDV):} Similar to the SCDV, we use Gaussian Mixture Model to cluster our final sense-disambiguated word vectors (obtain from earlier step) into $K$ partition of the words in the corpus.\footnote{Note this capital $K$ is very different from the small $k$ use in the word sense disambiguation. The value of $K$ depend of number of distinct high-level concepts covered in the corpus.} For each contextualized word vector $w_{i}$ of word $w$ $\in$ $V$, we created $K$ different word-cluster vectors of $d$ dimensions ($wc\vec{v}_{io}$) by weighting word’s embedding with sparse probability distribution for the $o^{th}$ cluster, i.e., P($c_{o}|w_{i}$). Then, we concatenate the $K$ word-cluster vectors ($wc\vec{v}_{io}$) and weight them with their inverse document frequency (idf($w_{i}$)) to form a contextualized word-topic vector ($w\vec{t}v_{i}$). The dimension of word-topic vector ($w\vec{t}v_{i}$) is $K$ $\times$ $d$. To obtain the final document embedding  $d\vec{v}_{D_{n}}$, we computed the average of the contextualized word-topic vectors $w\vec{t}v_{i}$ from the words and it's context as appearing in that document $D_{n}$. For more details on SCDV, refer to Algorithm $1$ in \citealp{mekala-etal-2017-scdv}. Figure \ref{tab:flowchart} shows flowchart of our approach (SCDV+BERT(cxtd)).

\section{Experimental Results}

We perform experiments with multi-class classification with data restriction settings, concept matching problem and sentence similarity task.

\paragraph{Datasets and Baselines:} We experimented on $6$ widely used classification dataset (in \emph{`English'}) whose statistics are shown in Appendix \S A Table \ref{tab:dataset_details}. We also validated our algorithm through the Concept Matching experiment on Concept-Project \cite{gong-etal-2018-document} dataset, where the task was to generate concepts from a given document corpus. We also perform experiments on the SemEval dataset (Y12 - Y16) involving 27 semantic textual similarity (STS) tasks from 2012 - 2016 \cite{agirre-etal-2012-semeval}. We represent our model as SCDV + BERT(ctxd), which is SCDV using multi sense-disambiguated contextual BERT word vector for document representation. We consider the following baselines for comparison: SCDV with single sense Word2Vec \cite{mikolov2013linguistic}, BERT(pr) \cite{devlin-etal-2019-bert} i.e. pre-trained BERT vectors averaging, BERT (fine-tune) \cite{sun2019fine} i.e. fine-tune BERT model, and Word2Vec (tfidf-weighted) i.e. a tf-idf weighted Word2Vec average. We also compare SCDV + BERT(ctxd) with an ablation representation obtain without corpus contextualization a.k.a, sense-disambiguation i.e. the value of WSD clustering parameter $k$ $=$ $1$ for all the words in the corpus. We referred this ablation representation as SCDV + BERT(weight-avg) in the paper. \footnote{Model hyperparameters are provided in Appendix \S A.} Furthermore, we also adopted the work of \citealp{ethayarajh-2019-contextual} which adjust Anisotropy (more uniformity) with our embeddings. We refer this representation as SCDV + BERT(ctxd) + Anisotropy. Following \citealp{ethayarajh-2019-contextual} recommendation, we use BERT last layer for word embeddings.

\paragraph{Full Data Setting:} Table \ref{tab:my-table} shows a comparison of accuracy across all the datasets. We observe that SCDV + BERT(ctxd) model outperforms the original SCDV+Word2Vec across all the datasets. To ablate the contribution of sense-disambiguation using BERT contextualization, we also compare SCDV + BERT(ctxd) result with SCDV + BERT(weight-avg). We also analyze the effect of reducing Anisotropy on SCDV + BERT(ctxd).\footnote{For top $1000$ words, the cosine similarity reduces from $0.5468$ to $0.3752$ after anisotropic adjustment.} 

\begin{figure*}[!htbp]
\centering
\includegraphics[height=1.1in]{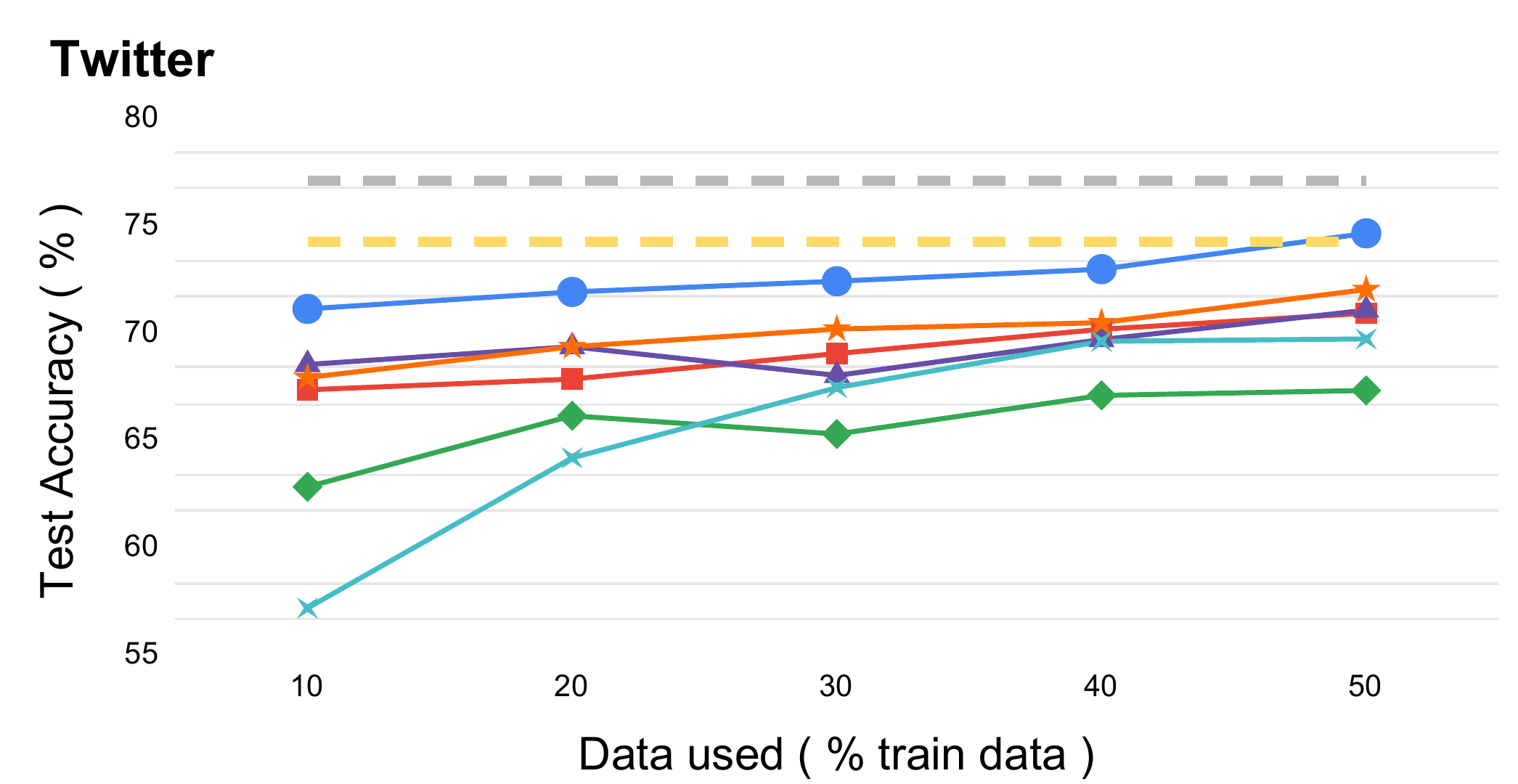}
\hspace{-0.5cm}\includegraphics[height=1.1in]{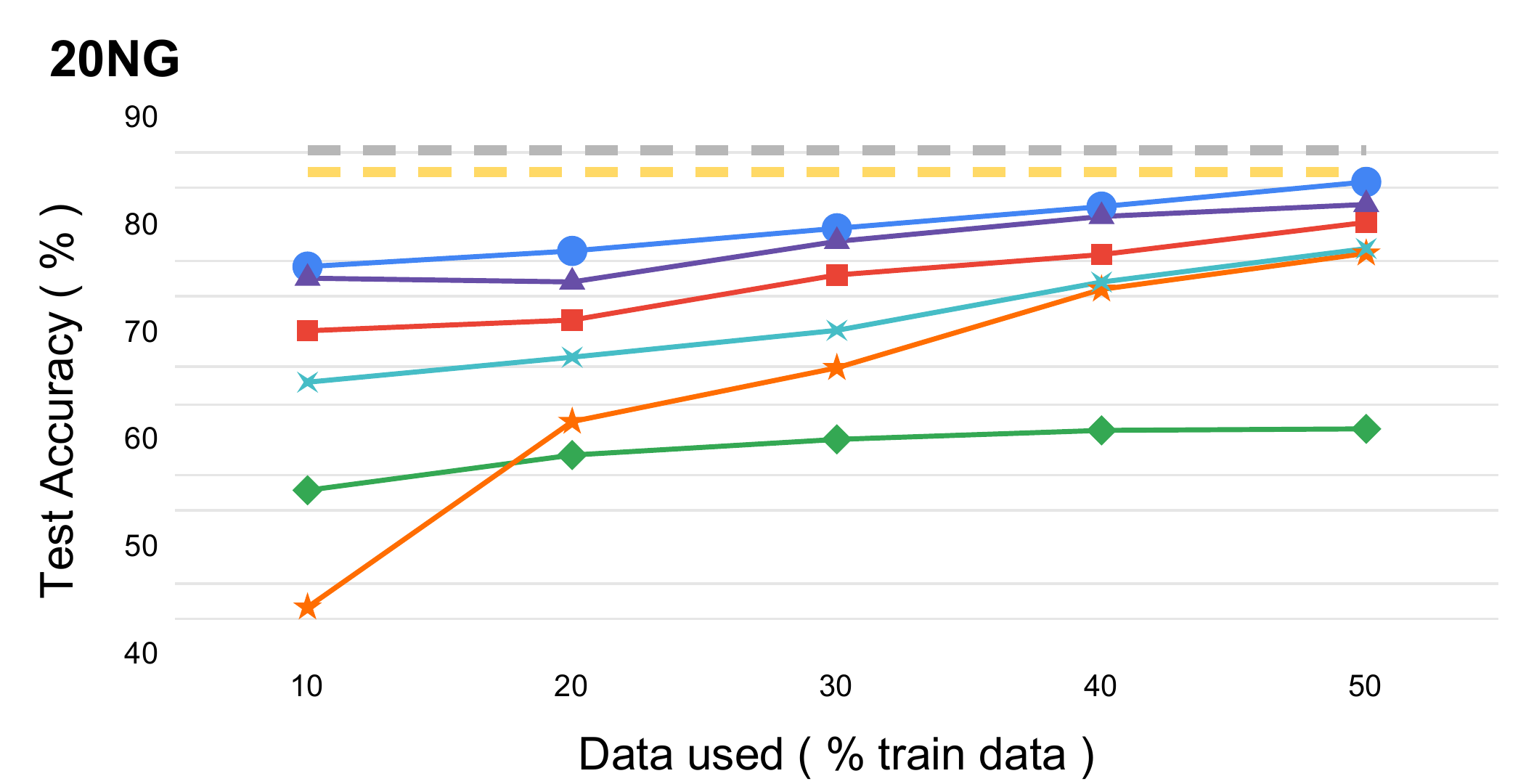}
\hspace{-0.5cm}\includegraphics[height=1.1in]{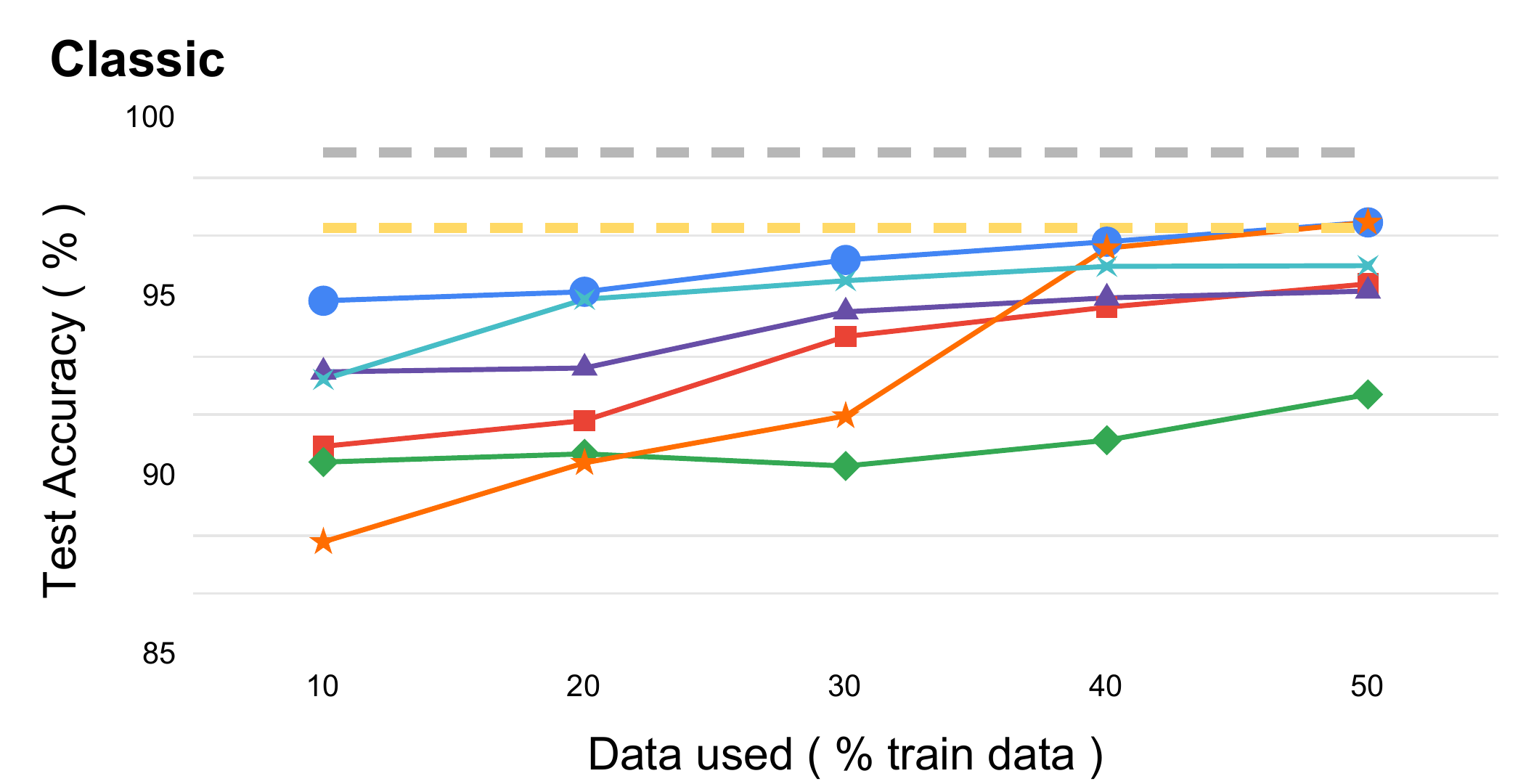}
\hspace{-0.5cm}\includegraphics[height=1.1in]{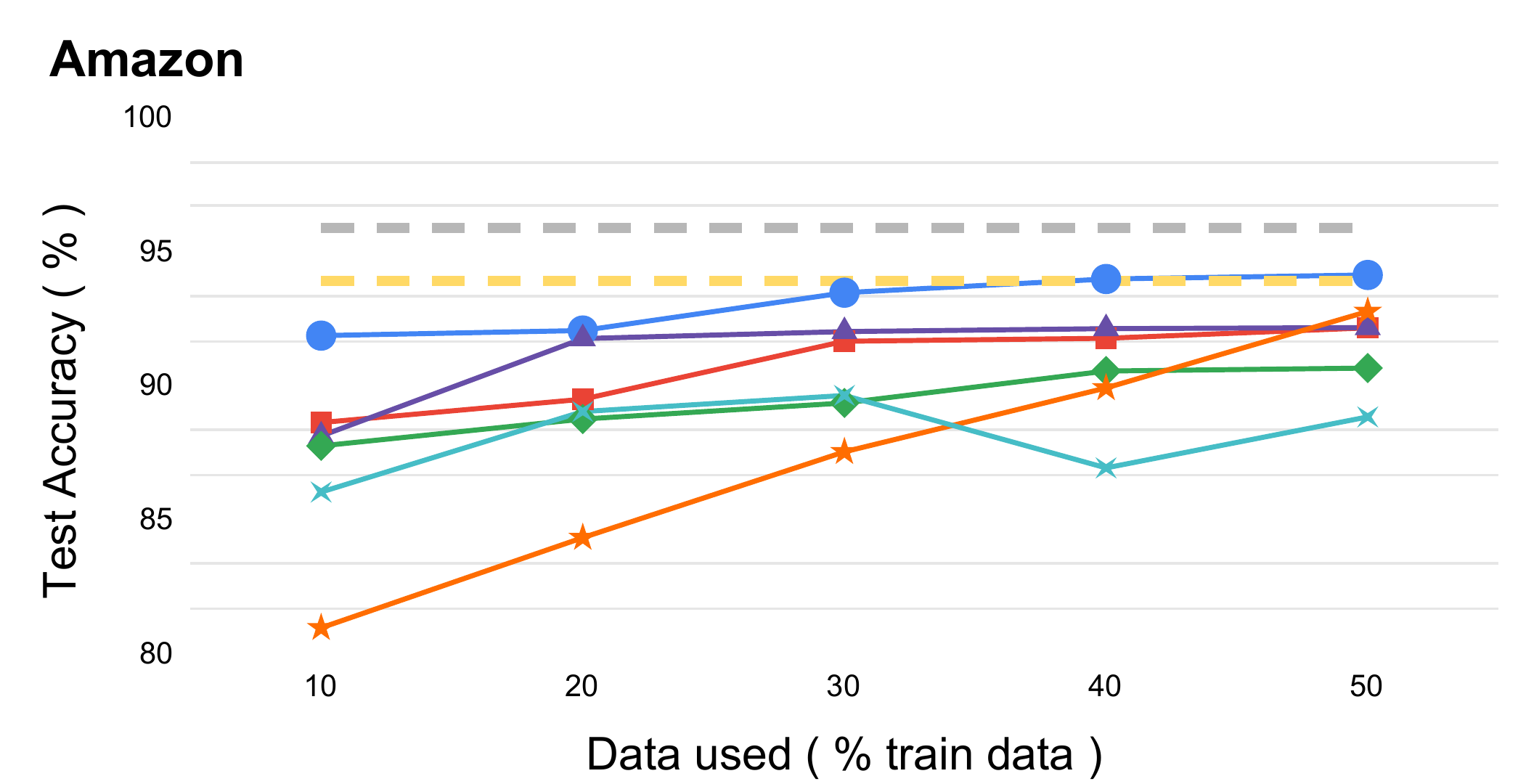}
\hspace{-0.5cm}\includegraphics[height=1.1in]{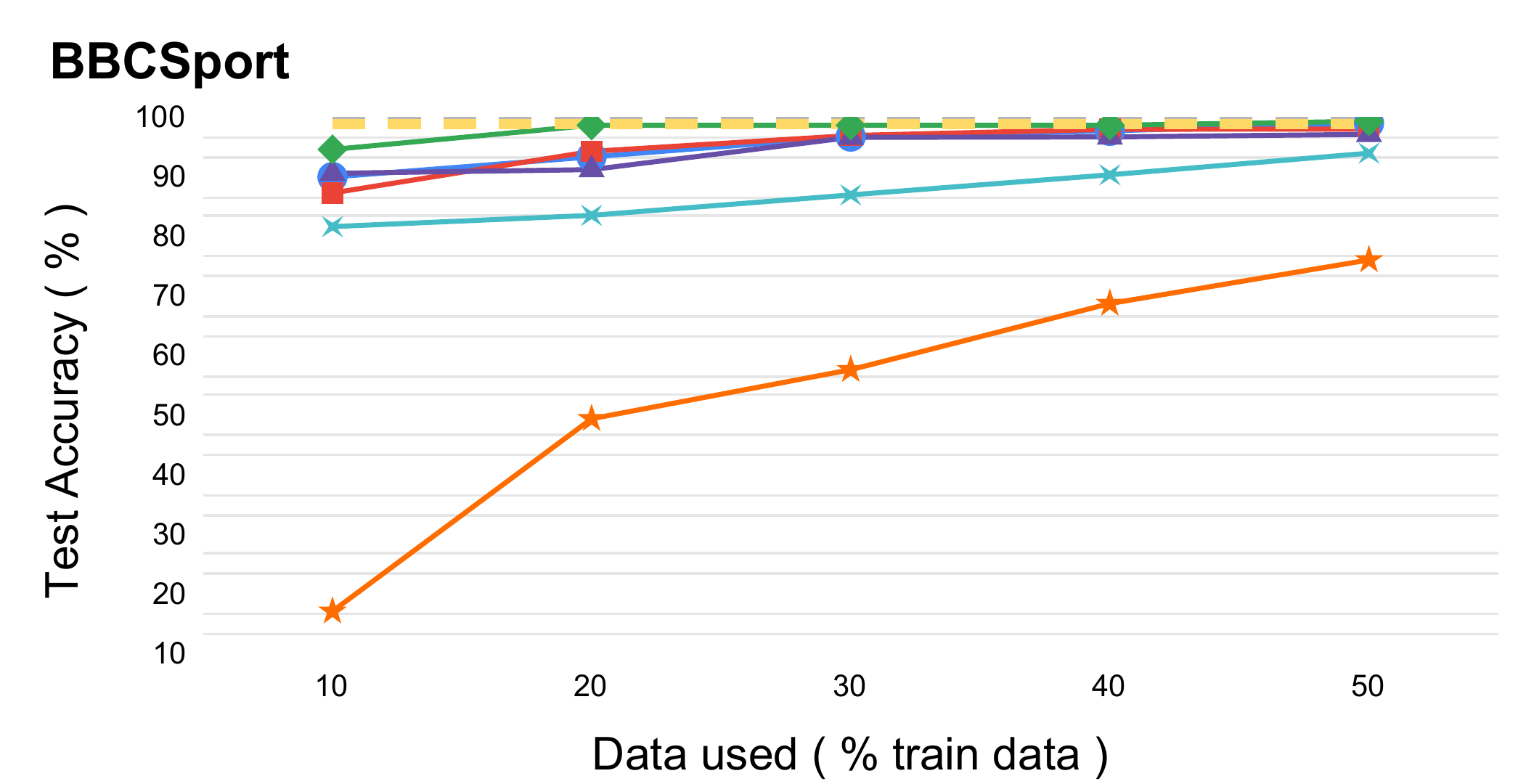}
\hspace{-0.5cm}\includegraphics[height=1.1in]{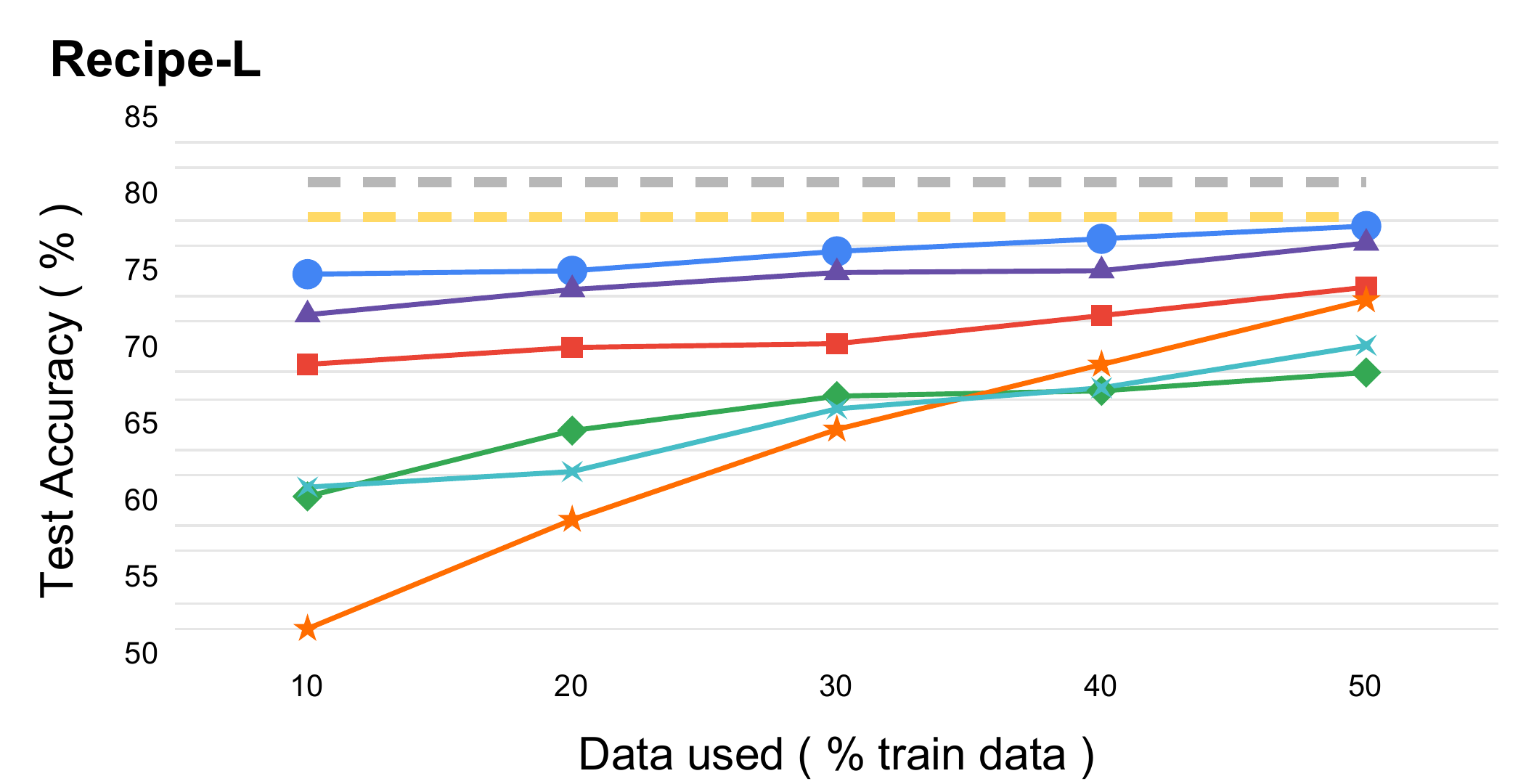}
\includegraphics[height=0.30in]{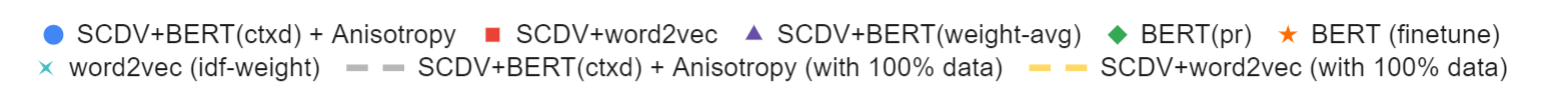}
\vspace{-0.5cm}
\caption{\small Embedding performance with limited training data i.e. semi-supervised setting. Standard deviation avg: 0.28 with range (0.03,0.81).}
\label{tab:semi-suervised}
\vspace{-1.5em}
\end{figure*}

\textit{Analysis}: Contextualized BERT (SCDV + BERT(ctxd)) performs better than the average BERT (SCDV + BERT(weight-avg)), which in turn performs competitively with original SCDV + Word2Vec. This indicates that the sense-disambiguated BERT based word vectors captures multiple meaning of word (better corpus contextualization) effectively. \footnote{Similar observation made by \citealt{gupta2020multisense} (SCDV-MS), an extension for SCDV with multi-sense word2Vec.}. Furthermore, reducing Anisotropy i.e. SCDV + BERT(ctxd) + Anisotropy, further boosts the performance of SCDV + BERT(ctxd). As expected, the BERT (fine-tune) performs the best on most datasets (except for Twitter and Amazon), where all versions of SCDV perform much better than pre-train BERT averaging i.e. BERT(pr). The good performance of BERT (fine-tune) is expected as fine-tuning modifies the model parameter (layer weights), producing task and domain-specific embedding, often impressed in the \emph{[CLS]} token representation. Note that SCDV + BERT(ctxd)+ Anisotropy are unsupervised embedding but had accuracy competitive (sometimes even better) to the supervised fine-tuned BERT models i.e. BERT (fine-tune). Thus, our approach could be used as an effective alternative to fine-tuning BERT for selected classification tasks.

\paragraph{Limited Data Setting:} BERT has produced state-of-the-art results in various NLP domains, but its use is restricted to the availability of labeled data, whereas with the help of pre-trained BERT and SCDV, our approach can also work with limited data i.e. semi-supervised setting,  requiring only sufficient enough labeled data to learn the downstream classifier. To test the effectiveness of our approach in low-data conditions, we ran the same multi-class classification experiment with $10\%$, $20\%$, $30\%$, $40\%$, and $50\%$ of the training data. See Figure \ref{tab:semi-suervised} for the results.

\textit{Analysis}: Contrast to earlier fully supervised settings, we observe that BERT (fine-tune) performed worst (except for Twitter data) due to limited training data. BERT (fine-tune) have a significant mean performance drop of greater than $33.5\%$ across datasets with limited data training of 10$\%$ data compare to full training with 100$\%$ data. Whereas, the performance of SCDV + BERT(ctxd) (and it's Anisotropic version i.e., SCDV + BERT(ctxd) + Anisotropic) remained comparable with the full data (i.e., $100\%$ data) setting, with the mean performance drop of just $\approx$ $7.2\%$ across datasets with the 10$\%$ data. \footnote{For some dataset such as BBCSport the performance drop was also $<$ $1\%$.} It also outperformed original SCDV + Word2Vec and SCDV+BERT(weighted-avg) on all datasets. \footnote{Except the BBCSport, where pre-trained BERT embeddings, i.e., BERT(pr), produced comparable results due to fewer multi-sense words as shown in Table 1 of Appendix \S A.} Under low data conditions, SCDV+BERT(weight-avg) also outperformed the original SCDV + Word2Vec due to added contextualization benefits of the pre-trained BERT word vector representation. Moreover, under significantly less data (10\% or 20\% data), even BERT(pr) and Word2Vec (tdf-idf weighted) performed significantly better than the BERT(fine-tune). Furthermore, we find that our method outperforms SCDV with complete training on a range of datasets while only employing limited 40$\%$ to 50$\%$ data. Overall, our unsupervised method significantly outperforms fine-tune BERT in the low data domain. We predict that fine-tuning BERT with little supervision is exceedingly difficult due to the needed learning of large-parameter space.

\paragraph{Few-Shot Setting:} We also experimented under few-shot conditions where the available data is too low for even training a classification (and obviously for BERT fine-tuning i.e. BERT(fine-tune) too). We implemented a K-shot N-way prototypical few-shot classifier where K is the number of samples used from each class and N is the number of classes in the dataset. We set the K values from $5$, $10$, $15$, and $20$ for our setting, and N is equal to numbers of class labels (see Appendix \S A). New examples are assigned the label using the nearest neighbor approach (KNN algorithm with K = 1), i.e., the label of closest averaged prototypical class point. 

\begin{figure}[!htbp]
\centering
\includegraphics[height=1.35in]{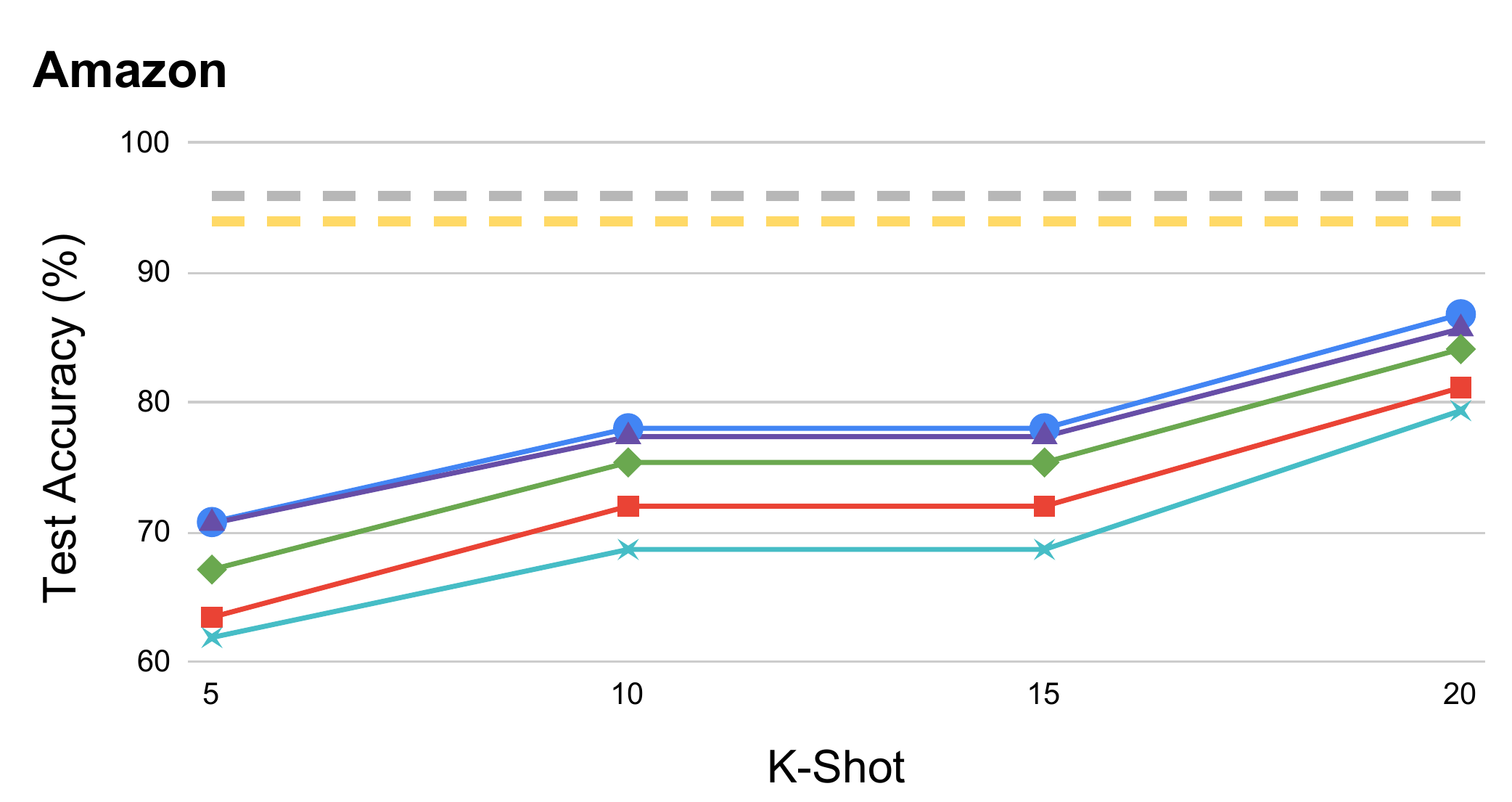}
\includegraphics[height=1.35in]{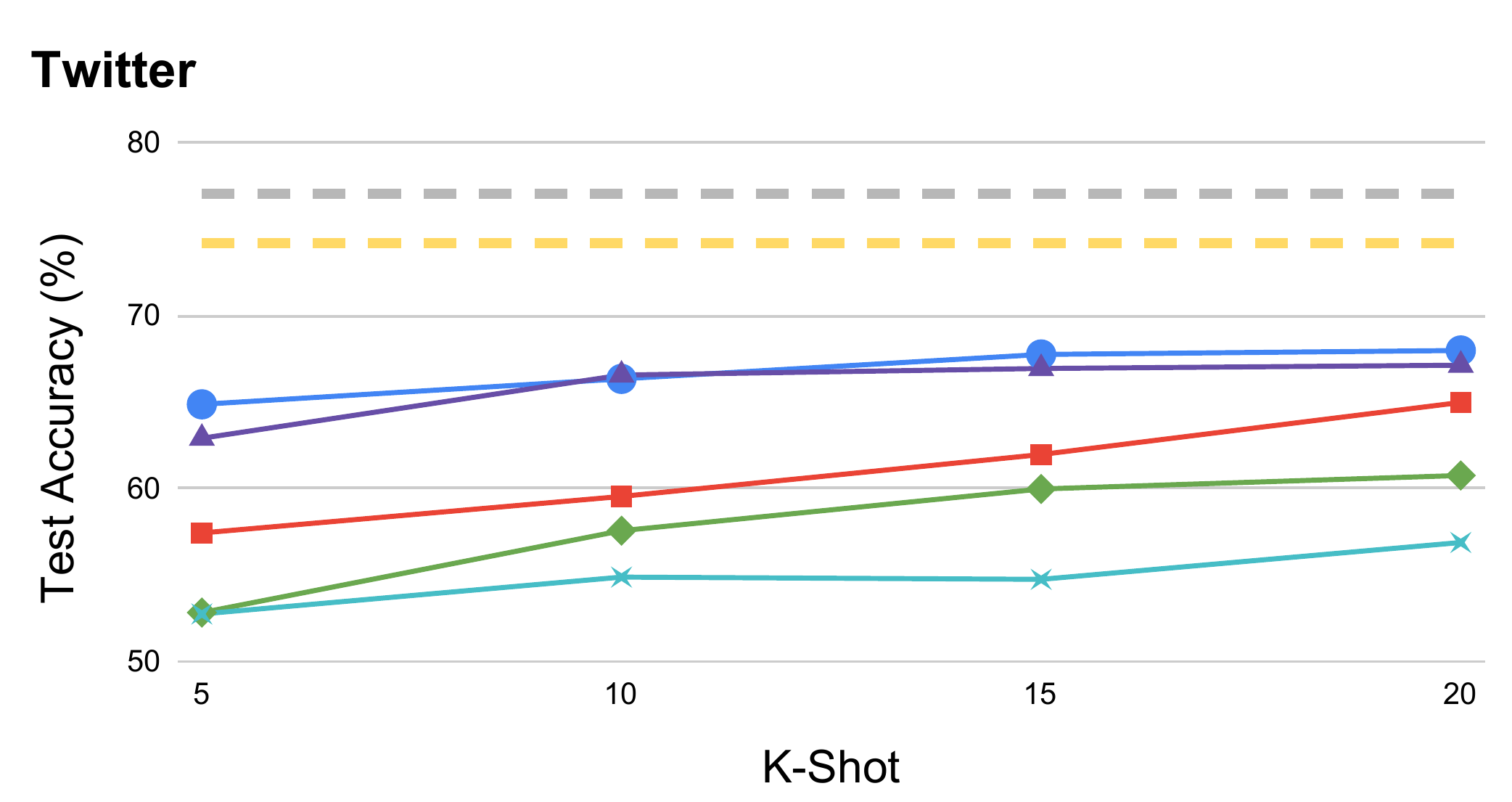}
\includegraphics[height=0.4in]{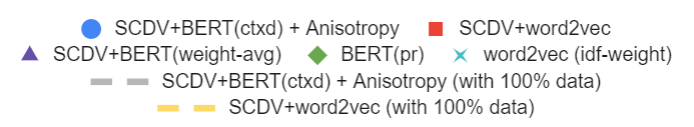}
\vspace{-0.5em}
\caption{\small Results for the few-shot experiment. Reported number are mean of $5$ random runs with having standard deviation average: $0.14$ with range of (0.09, 0.43)}
\label{tab:few-shot}
\vspace{-1.25em}
\end{figure}

\textit{Analysis}: We see in Figure \ref{tab:few-shot} that SCDV + BERT(ctxd) and SCDV+BERT(weight-avg) outperform all other methods by a significant margin ($>11\%$ mean across datasets). SCDV+BERT(ctxd) for most of the time marginally leads SCDV+BERT(weighted-avg), showing that contextualization also helps in the few shots settings. In contrast to earlier results adjusting anisotropy over SCDV + BERT(ctxd) only improve performance marginally. Thus, our approach is effective in scenarios where classifier training is impossible.

\paragraph{Concept Matching:}  The task is to link the concept with the relevant projects. Concept Matching dataset includes $537$ pairs of projects
and concepts involving $53$ unique concepts from the Next Generation Science Standards3 (NGSS) and $230$ unique projects from Science Buddies. We compare  cosine similarity between our method with the TF-IDF-weighted vectors, SCDV + Word2Vec, InferSent \cite{conneau-EtAl:2017:EMNLP2017} and pre-trained BERT(pr) baselines. From table \ref{tab:concept-matching}, we observed that our algorithm (SCDV + BERT(ctxd) + Anisotropy) outperformed pre-trained BERT and SCDV + Word2Vec by 5.2$\%$ and 4.6$\%$ on F1 and accuracy respectively on the Concept-Project \cite{gong-etal-2018-document} dataset. 

\begin{table}[!htbp]
\small
\centering
\begin{tabular}{ c|c|c } 
 \toprule
 \bf Embedding & \bf Accuracy & \bf F1  \\
 \midrule
%  tf-idf & 53.8 & 70.0  \\ 
TF-IDF & 53.8 & 70.0 \\
InferSent & 54.0 & 70.1 \\
BERT(pr) & 54.8$_{(0.2)}$ & 70.6$_{(0.3)}$  \\ 
SCDV + Word2Vec & 53.7$_{(0.1)}$ & 70.0$_{(0.1)}$ \\ \hdashline
SCDV + BERT(ctxd) &  57.1$_{(0.2)}$ &  73.8$_{(0.2)}$ \\
SCDV + BERT(ctxd)  & \bf 58.9$_{(0.1)}$ & \bf 74.6$_{(0.1)}$ \\ 
+ Anisotropy & & \\
\bottomrule
\end{tabular}
\caption{\small Embedding performance on Concept-Marching dataset. Bold represents best performance. Baselines are taken from  \citealp{zhang-danescu-niculescu-mizil-2020-balancing}.}
\label{tab:concept-matching}
\vspace{-1.25em}
\end{table}

\paragraph{Sentence Similarity Task:}  The objectives of these tasks are to predict the similarity between two sentences. Performance is assessed by computing the Pearson correlation \cite{freedman2007statistics} between machine-assigned semantic similarity scores and ground truth. SCDV + BERT(ctxd) + Anisotropy substantially outperform several other baselines as demonstrated in the table \ref{tab:STS}.

\begin{table}[!htbp]
\small
\centering
\setlength\tabcolsep{3.5pt} 
\begin{tabular}{ c|c|c|c|c|c|c } 
 \toprule
 \bf Embedding & \bf Y12 & \bf Y13 & \bf Y14 & \bf Y15 & \bf Y16 & \bf Avg. \\
 \midrule
 ELMO orig+all & 55 & 51 & 63 & 69 & 64 & 60.4 \\
 ELMO orig+top & 54 & 49 & 62 & 67 & 63 & 59 \\
 BERT(pr) Avg. & 53 & 67 & 62 & 73 & 67 & 64.4 \\
 USE & 65 &\bf 68 & 64 & 77 & 73 & 69.4 \\
p-mean & 54 & 52 & 63 & 66 & 67 & 60.4 \\
 fastText & 58 & 58 & 65 & 68 & 64 & 62.6 \\
 Skip Thoughts & 41 & 29 & 40 & 46 & 52 & 41.6 \\ 
 InferSent & 61 & 56 & 68 & 71 & 77 & 66.6 \\ 
 PSIF + PSL & 65.7 & 64.0 & 74.8 & 77.3 & 73.7 & 71.1 \\
 u-SIF + PSL & 65.8 & 65.2 & 75.9 &  77.6 & 72.3 & 71.4 \\ 
 SCDV + WordVec &  64.1 & 63.9 &  73.0 & 76.9 &\bf 77.3 & 71.0 \\ 
 \hdashline
 SCDV + BERT(ctxd) &  64.7 &\bf 64.0 &  75.4 & 77.1 & 73.3 & 70.9 \\
 SCDV + BERT(ctxd)  &  \bf 66.8 & 64.1 &  \bf 77.3 &\bf 78.0 &  74.6 &\bf 72.2 \\
 + Anisotropy & & & & & & \\
 \bottomrule
\end{tabular}
\vspace{-0.5em}
\caption{\small Embedding performance on Semantic Textual Similarity task (STS) with several embeddings for each year with overall average (avg.). Bold represent best performance. Baselines are taken from \citealp{gupta2020psif}.}
\label{tab:STS}
\vspace{-1.5em}
\end{table}

\section{Comparison with Related Works}

The closest work to our paper is SCDV by \citealp{mekala-etal-2017-scdv} that extend BoWV by \citealp{gupta-etal-2016-product} using an overlapping clustering technique and direct idf weighting of word vectors. Recently \citealp{gupta2020multisense} extended SCDV to SCDV-MS via utilising the multi-sense embeddings obtained via using AdaGram~\cite{pmlr-v51-bartunov16} over WordVectors ~\cite{mikolov2013linguistic}. Our idea is similar to SCDV-MS; however, it utilises pre-train BERT contextual embedding as word embedding, a.k.a a more robust sense disambiguated aware embedding \cite{mekala-etal-2020-meta}. Thus, our approach (SCDV + BERT(ctxd)) uses contextual word vectors as the foundational block for document representation.

\section{Conclusion and Future Work}
\label{conclusion}
In this paper, we enhance sparse document representation (SCDV) with pre-trained BERT contextualization and propose SCDV+BERT(ctxd).  We showed that one could effectively utilize the BERT contextualization for word-sense disambiguation. Our approach outperforms other unsupervised approaches in the full data regime. Our approach is also very successful for low data regime, outperforming the standard model with roughly half the training data required and few shot settings, where fine-tuning of model fails. 

\section{Acknowledgement}
We appreciate the Utah NLP group members' valuable suggestions at various phases of the project, as well as the reviewers' helpful remarks.

\bibliography{anthology,custom}
\bibliographystyle{acl_natbib}

\appendix

\section{Hyper parameter Details}
\label{sec:hyperparameter}

\begin{figure*}
\centering
\includegraphics[width=\columnwidth]{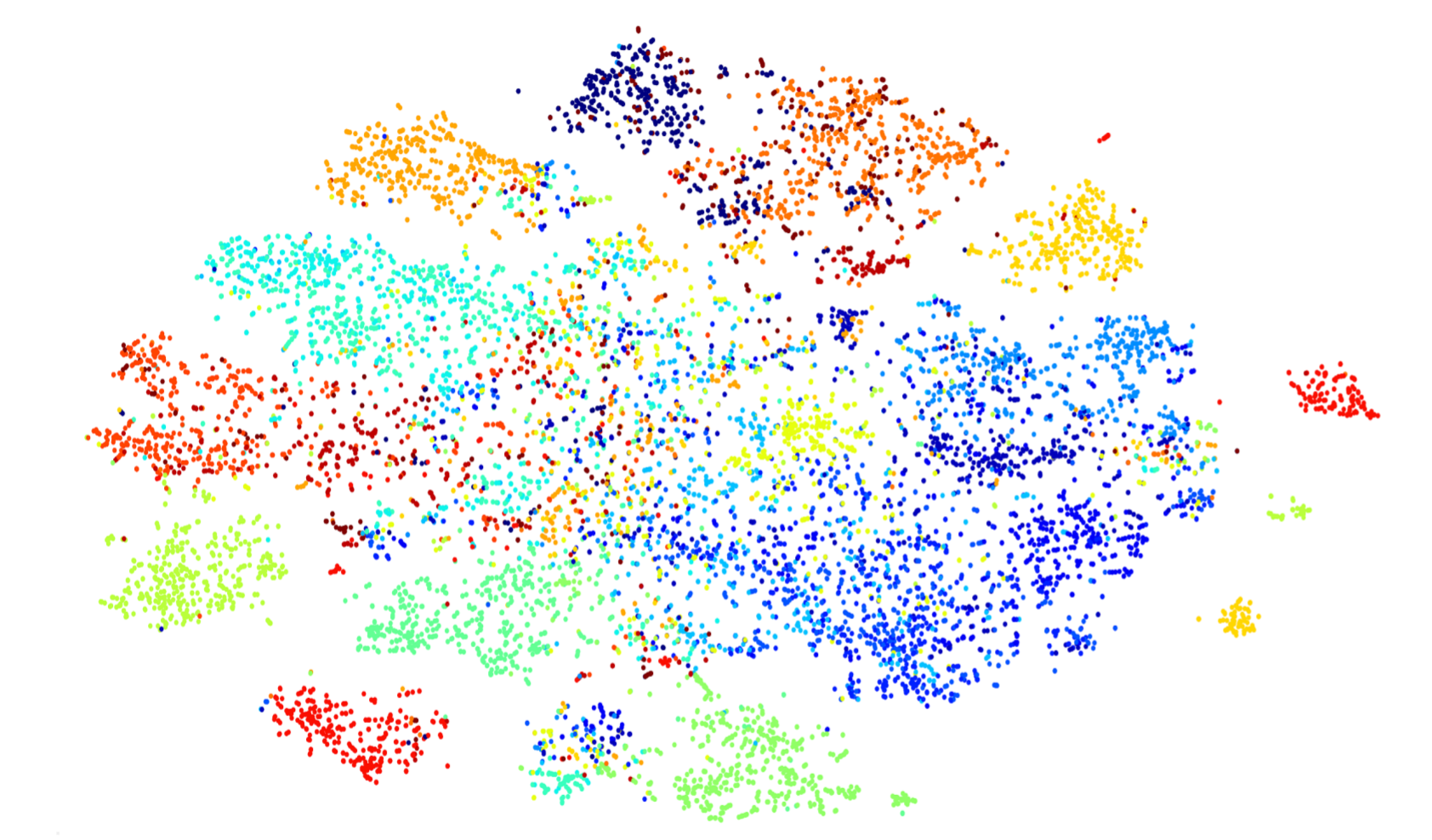}
\includegraphics[width=\columnwidth]{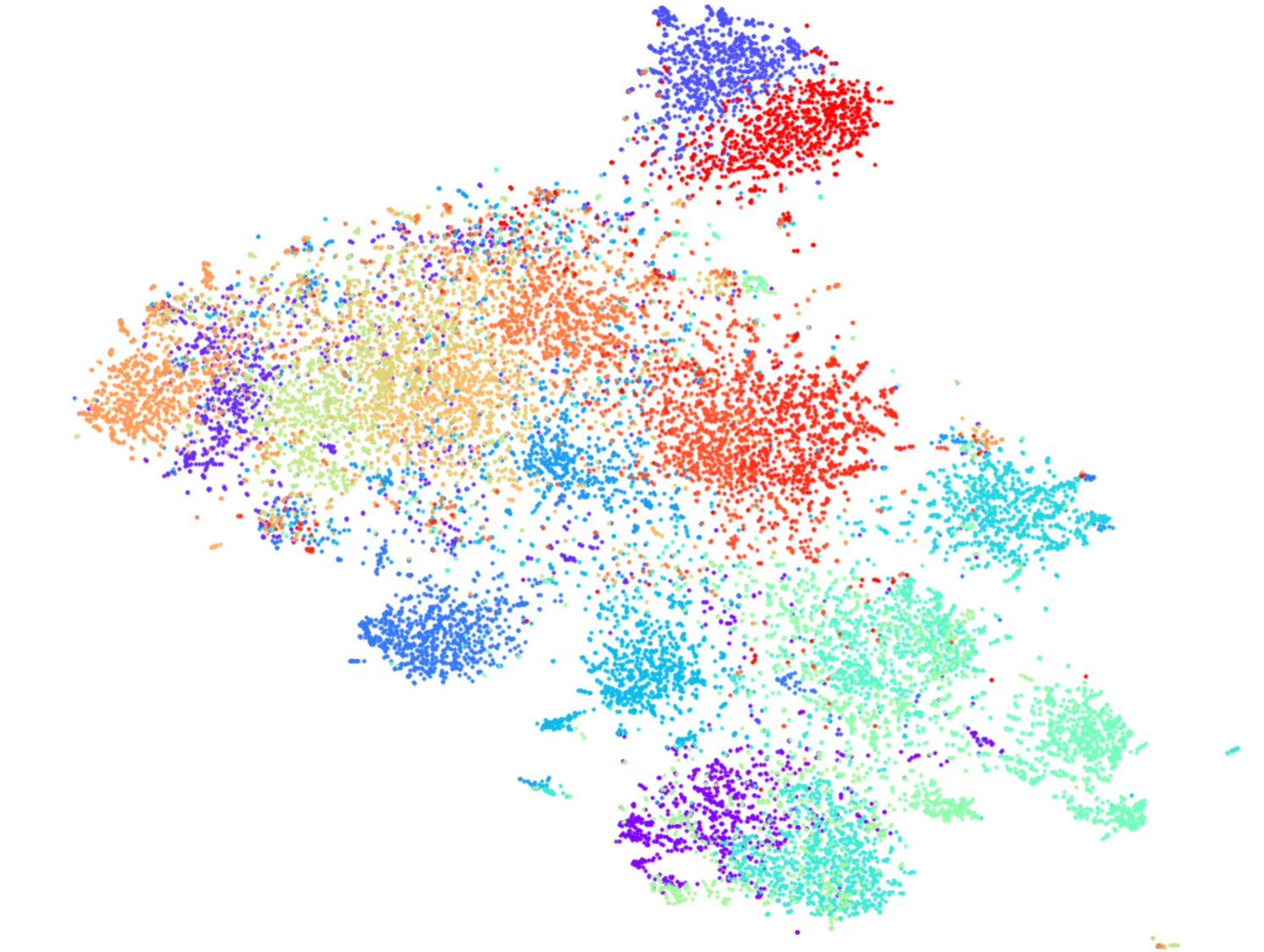}
\vspace{-0.25em}
\caption{\small t-SNE plots: SCDV+WordVec(left) and SCDV+BERT(ctxd) + Anisotropy(right). Clearly, better class separation for SCDV+BERT(ctxd) + Anisotropy.}
\label{fig:t-sne_plots}
 \vspace{-1.25em}
\end{figure*}

We obtain the word embeddings using BERT-base-uncased pre-trained model and use k-means for contextual clustering for given word. For simplicity we used similarity threshold ($\tau$) of $0.8$ for all words in all the datasets \footnote{See datasets details in Table \ref{tab:dataset_details}. These datasets have been the subject of a number of previous research studies.} which lead to multiple polysemous word representation for each word, the distribution for the same is shown in Table \ref{tab:polysemous_details}. For SCDV, we set the dimension of word embeddings to $200$ and the number of mixture components in GMM between $30 - 90$ (dataset dependent) as shown in Table \ref{tab:number_clusters}. For the GMM we ensure that all mixture components share the same \emph{tied co-variance} matrix. Sparsifying the document vectors further as propose in SCDV leads to only marginal performance gain (statistically insignificant), so we skip that step for our experiments. We used LinearSVM for multi-class classification during fully supervised and semi-supervised settings and prototypical networks for few-shot setting. The choice of the classifier was the same in all baselines and the proposed model to maintain uniformity. We used 5-fold cross-validation on the F1 score to tune parameter C of LinearSVM. In semi-supervised setting the example are added in incremental setting for fair comparison. We also repeated the experiment $10$ times (varying random set selection seed) and consider mean as our final performance. 

\begin{table}[!htbp]
\small
    \centering
    \begin{tabular}{ c|c|c|c } 
     \toprule
     \bf Dataset & \bf Train \bf Size & \bf Test Size &\bf $\#$Label \\
     \midrule
     20NG & 11314 & 7358 & 20 \\ 
     Amazon & 5600 & 2400 & 4 \\ 
     Twitter & 2180 & 935 & 3 \\ 
     BBCSport & 516 & 221 & 5\\ 
     Classic & 4966 & 2129 & 4 \\ 
     Recipe-l & 27842 & 11932 & 20\\
     \bottomrule
    \end{tabular}
    \caption{Dataset Statistics.}
    \label{tab:dataset_details}
    \vspace{-1.25em}
\end{table}

\textbf{STS Task Details: } For the STS task, the gold score is a continuous valued similarity score on a scale from $0$ to $5$, with $0$ indicating that the semantics of the sentences are completely independent and $5$ signifying semantic equivalence is computed.

\begin{table}[!htbp]
\centering
\small
    \begin{tabular}{ c|c|c|c } 
     \toprule
     \bf Dataset & \bf k=1 & \bf k=2 & \bf k$\geq$3   \\
     \midrule
     20NG & 80.29 & 13.58 & 6.23  \\ 
     Amazon & 76.12 & 17.68 & 6.20  \\ 
     Twitter & 80.79 & 15.60 & 3.61 \\ 
     BBCSport & 87.29 & 11.56 & 1.15 \\ 
     Classic & 73.63 & 17.01 & 9.36 \\ 
     Recipe-l & 67.11 & 13.98 & 18.91 \\
     \bottomrule
    \end{tabular}
    \vspace{-0.25em}
    \caption{\small Distribution (in \%) of vocabulary as it's disambiguated into k = 1, 2 and $\geq3$ polysemous words.}
    \label{tab:polysemous_details}
    \vspace{-1.00em}
\end{table}

\begin{table}[!htbp]
\centering
\small 
    \begin{tabular}{c|c|c|c|c|c|c} 
     \toprule
     \bf Dataset ($\%$) &\bf 10 &\bf 20 &\bf 30 &\bf 40 &\bf 50 &\bf 100 \\
     \midrule
     20NG & 45 & 45 & 60 & 60 & 60 & 60  \\ 
     Amazon & 30 & 30 & 30 & 30 & 30 & 30  \\ 
     Twitter & 30 & 45 & 45 & 45 & 45 & 45 \\ 
     BBCSport & 60 & 60 & 75 & 75 & 75 & 90  \\ 
     Classic & 30 & 30 & 30 & 30 & 30 & 30 \\ 
     Recipe-l & 30 & 30 & 30 & 30 & 30 & 30  \\
     \bottomrule
    \end{tabular}
    \vspace{-0.25em}
    \caption{Number of mixture components in GMM used in various experiments. \footnote{For Few-Shot setting the number of components were taken same as 10\% data}}
    \label{tab:number_clusters}
    \vspace{-1.25em}
\end{table}

\section{Word Sense Disambiguation Examples}
\label{sec:wsd_examples}

Table \ref{tab:dataset_similarity_score} shows word sense disambiguation for few polysemous words from 20NewsGroup dataset along with the  cosine similarity between BERT embedding with different context usage. We use threshold of $0.8$ for sense cluster disambiguation. Figure \ref{fig:t-sne_plots} shows tha t-sne plots for SCDV+WordVec(left) and SCDV+BERT(ctxd) + Anisotropy(right) embeddings. Clearly, we see much better class separation for SCDV+BERT(ctxd) + Anisotropy than  SCDV+WordVec. We can alse the anisotropic reduction conical effect.

\begin{table}[!htbp]
\small
\centering
\begin{tabular}{ c|c|c } 
\toprule
\bf  Word & \bf Sentence & \bf Score  \\
\midrule
Subject & The math \textbf{subject1} is difficult \\ 
& He sent the mail without \textbf{subject2} & 0.71 \\
\midrule
Apple & The stocks of \textbf{Apple1} have increased \\ 
& I eat an \textbf{apple2} everyday& 0.67\\ 
\midrule
Unit & Metre is \textbf{unit1} of Distance \\
& He is in 1st \textbf{unit2} & 0.78 \\ 
\bottomrule
\end{tabular}
\caption{\footnotesize Word Sense Disambiguation. Here, score represent the  cosine similarity between BERT embedding of the \textbf{bold} subject word.}
\label{tab:dataset_similarity_score}
\end{table}

\section{Other Related Work}
\label{sec:other_related_work}

\citealp{Levy2014NeuralWE} used unweighted averaging of word vectors, \citealp{singh-mukerjee-2015-words} proposed tfidf-weighted averaging of word vectors, \citealp{socher-etal-2013-recursive} proposed a recursive neural network defined over a parse tree with supervised training. \citealp{10.5555/3044805.3045025} proposed PV-DM and PV-DBoW models which treat each sentence as a shared global latent vector. Other approaches use seq2seq models such as RNN~\cite{inproceedings} and LSTM~\cite{article} which can handle long term dependency. \citealp{wieting2017revisiting} proposed a neural network model which optimizes the word embeddings based on the cosine similarity. Recent deep contextual word embeddings such as ELMo \cite{peters2018deep}, USE \cite{cer2018universal} and BERT \cite{devlin-etal-2019-bert}, which capture the word context, outperform all earlier approaches. There are also other topic based modeling approaches such as LDA \cite{Chen2017}, weight-BoC \cite{Kim2017BagofconceptsCD}, TWE \cite{10.5555/2886521.2886657} , NTSG \cite{Liu2015LearningCW}, w2v-LDA \cite{nguyen-etal-2015-improving}, etc, as explored in SCDV \cite{mekala-etal-2017-scdv}. Recently, \cite{houlsby2019parameterefficient} propose a parameter efficient fine-tuning of transformer model (e.g. BERT) via selective training of top layers for transfer learning. Compare to this our approach also work doesn't do any fine-tuning totally unsupervised and even work in semi-supervised and few-shot setting, where training a classifier is difficult.

\end{document}